\documentclass[aip,reprint]{revtex4-1}

\usepackage{upgreek} 
\usepackage{lipsum} 
\usepackage{graphicx}
\usepackage{wasysym} 
\usepackage{amssymb} 
\usepackage[monochrome]{xcolor} 

\draft 

\begin{document}


\title{Miniature magneto-oscillatory wireless sensor for magnetic field and gradient measurements} 

\author{F. Fischer}
\affiliation{Division of Smart Technologies for Tumor Therapy, German Cancer Research Center (DKFZ) Site Dresden, 01307 Dresden, Germany}
\affiliation{Faculty of Engineering Sciences, University of Heidelberg, 69120 Heidelberg, Germany}

\author{M. Jeong}
\affiliation{Institute of Physical Chemistry, University of Stuttgart, 70569 Stuttgart, Germany}

\author{T. Qiu}
\email[]{tian.qiu@dkfz.de}
\affiliation{Division of Smart Technologies for Tumor Therapy, German Cancer Research Center (DKFZ) Site Dresden, 01307 Dresden, Germany}
\affiliation{Faculty of Medicine Carl Gustav Carus, Dresden University of Technology, 01307 Dresden, Germany}
\affiliation{Faculty of Electrical and Computer Engineering, Dresden University of Technology, 01187 Dresden, Germany}

\date{\today}

\begin{abstract} 
Magneto-oscillatory devices have been recently developed as very potent wireless miniature position trackers and sensors with an exceptional accuracy and sensing distance for surgical and robotic applications.
However, it is still unclear to which extend a mechanically resonating sub-millimeter magnet interacts with external magnetic fields or gradients, which induce frequency shifts of sub-mHz to several Hz and therefore affect the sensing accuracy.
Here, we investigate this effect experimentally on a cantilever-based magneto-oscillatory wireless sensor (MOWS) and build an analytical model concerning magnetic and mechanical interactions.
The millimeter-scale MOWS is capable to detect magnetic fields with sub-$\upmu$T resolution to at least ± 5 mT, and simultaneously detects magnetic field gradients with a resolution of 65 $\upmu$T/m to at least ± 50 mT/m.
The magnetic field sensitivity allows direct calculation of mechanical device properties, and by rotation, individual contributions of the magnetic field and gradient can be analyzed.
The derived model is general and can be applied to other magneto-oscillatory systems interacting with magnetic environments.
\end{abstract}

\maketitle 

\begin{figure*}
    \centering
    \includegraphics{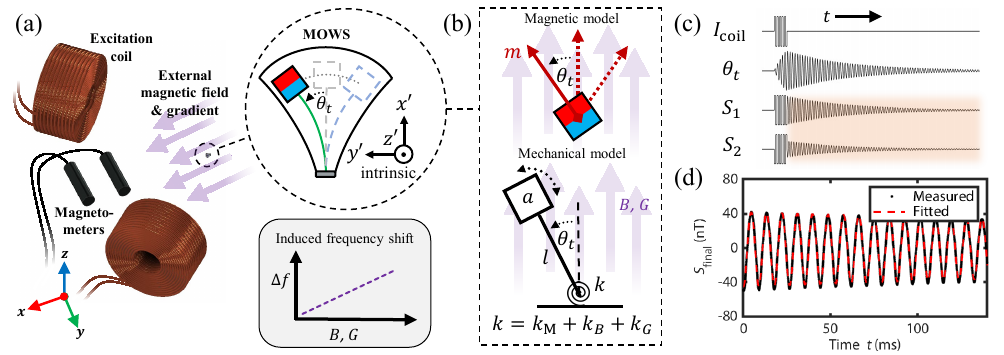}
    \caption{\label{fig:1} 
    (a) Schematic overview of the system to scale with an enlarged side view of the MOWS in its intrinsic coordinate system.
    The frequency of the magnets oscillation depends on the magnetic field $B$ and gradient $G$ in the main direction $x'$. 
    (b) Physical sub-models of the cantilever oscillator with the deflection angle $\uptheta_t$.
    (c) Time sequence of a signal acquisition with an excitation coil current $I_\mathrm{coil}$ to excite the magnet's oscillation.
    Two magnetometers $S_1$ and $S_2$ pick up the MOWS signal which is used for evaluation, shown as a shaded area.
    (d) Exemplary measured MOWS signal recorded at an effective distance of 6.75 cm and its corresponding signal fit.
    }
\end{figure*}
Many state-of-the-art sensors for physical parameters, often applied in skin-interfaced devices, which measure for example temperature, deformation, and magnetic fields, rely on complex electrically-responsive structures for energy and information transfer to achieve high precision and fast rates \cite{li2020,wang2018,kuzubasoglu2020,garcia2017}.
Although they achieve high sensitivities and can be embedded in wearable devices, their often-required wires inherently limit the field of operation, especially for surgical or robotic applications, where additional wires are obstructive.
Wireless solutions are typically based on radio-frequency (RF) communication and require either on-board batteries, energy-receiving coils or passive energy harvesting modules\cite{cui2019,min2018,huang2016}, which all occupy a large portion of the device.
Additionally, RF signals are strongly attenuated by human tissues, water or various metals \cite{werber2006,qureshi2016}.

Magnetic field sensors have been explored using the Hall-effect, various magnetoresistance and magnetoimpedance mechanisms\cite{khan2021,blanc2009,fernandez2015}, miniaturized fluxgate sensors \cite{wei2021,wang2022} or various MEMS devices \cite{herrera2016}.
Even though they show good performance in their working range, only few are suitable for completely untethered operation due to their electrical connectivity demands. 
Wireless low-power Hall sensors can be used at larger distances but typically occupy large volumes beyond 500 mm³ for power circuitry \cite{hu2022,sun2016}.
Fiber-optical methods have been used for a wireless read-out, however at negligible distances, unsuitable for practicable wireless operation \cite{long2016}.
An unpowered miniature wireless sensor using a magnetically-coupled mirror, which reflects an incoming laser beam, has been demonstrated to work at 1 m distance, however it can only detect strong fields $>$1 mT with an accuracy of 0.1 mT, requires a large photosensor array and does not allow reorientation \cite{vasquez2007}.

Recently, we reported a small-scale magneto-oscillatory localization (SMOL) device, which couples a magnetic moment to a cantilever, creating a mechanically resonating system which can be wirelessly excited with a magnetic field and read out by external magnetometers \cite{fischer2024}.
Due to resonance, the magnetic signal can be efficiently filtered to achieve a high signal-to-noise ratio.
A 3$\times$4$\times$4 mm³ cantilever-based device \cite{fischer2024} can be localized with high accuracy and refresh rates at large distances above 10 cm, and Gleich et al.\cite{gleich2023} reported a design with torsional oscillating magnetic spheres to achieve compact trackers below 1 mm³.
Besides for localization, they can be additionally used as wireless sensors using frequency-encoding properties for efficient sensing of e.g.~temperature and pressure \cite{gleich2023}, making them a very promising tool.

We developed magneto-oscillatory wireless sensors (MOWS) that can sense the mechanical properties of hydrogels and soft tissues\cite{fischer2022,fischer2023}.
Due to the magneto-mechanical interaction, external magnetic fields, such as the earth's magnetic field or ferromagnetic tools, can heavily influence the resonance frequency of the system, making the frequency-based sensing reliable 
on field-free, static, or magnetically well-defined environments.
The detailed analytical model describing the interaction between an oscillatory magnetic moment and an external field or field gradient, specifically for cantilever-based sensors, is still missing.
This model is not only critical for the precise measurements of above-mentioned physical parameters on MOWS, but can also be conversely utilized for the magnetic field and gradient sensing, and even mapping, in locations where a bulky wired sensor is not applicable.

In this work, we report the theoretical background and provide experimental verification for the magnetic field dependent behavior of a cantilever-based MOWS.
The derived theory reveals that not only the magnetic field but also the magnetic gradient act as independent virtual springs parallel to the mechanical spring.
Linear and non-linear frequency shift regions are evaluated, and the magnetic environment is mapped by rotation, which leads to precise information about the external magnetic fields and gradients by decomposition of both contributions. 
The frequency sensitivity for weak fields, additionally, allows determination of the system's mechanical properties.
The derived analytic equations are generally applicable to other sensitive magneto-oscillatory devices operated in magnetic environments.
Fundamental equations and supporting illustrations for a complete derivation of shown relationships are presented in the supplementary material.

An overview of the system is presented in Fig.~\ref{fig:1}(a). It consists of two wired magnetometers, two excitation coils and a cantilever-based MOWS (a detailed side view in the intrinsic coordinate system is shown as the inset).
Under the application of an external magnetic field of magnitude $B$ or a magnetic field gradient of magnitude $G$ in the MOWS main direction $x'$, the in-plane dynamic motion of the magnet is influenced, resulting in a frequency shift.
For derivation of the analytical solutions, the system is divided into two simplified magnetic and mechanical subsystems, as shown in Fig.~\ref{fig:1}(b).
The mechanical system is approximated by a torsional oscillator comprising a spring constant $k$, a point mass $a$ and a rigid beam of length $l$, while the permanent magnet is approximated as a dipole with a magnetic moment $m$.
The deflection angle $\uptheta_t$ over time $t$ of the mass, see Fig.~\ref{fig:1}(a) and (b), is described by an underdamped harmonic oscillation with a maximum deflection angle $\uptheta_{t,\mathrm{max}}$, a resonance frequency $f$ and a damping coefficient $\updelta$.
The field $B$ and the field gradient $G$ induce independent torques on the magnet, such that the overall spring constant $k$ is a sum of the mechanical-, magnetic field-, and magnetic gradient spring constants 
\begin{equation}\label{eq:k}
    k = k_\mathrm{M} + k_B + k_G,
\end{equation}
which is discussed below in detail to explain the frequency-shift phenomenon.
\begin{figure*}
    \centering
    \includegraphics{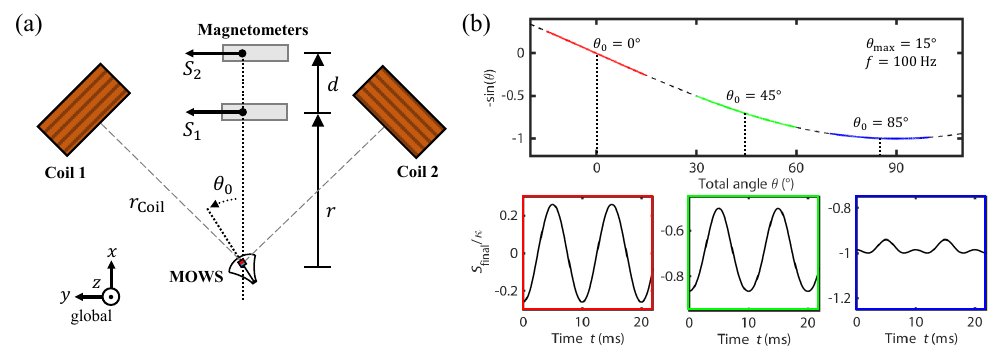}
    \caption{\label{fig:2} 
    (a) Schematic experimental setup from the top view in global coordinates and two coils at 45° angles for excitation of the MOWS.
    (b) Examples of magnetic signals $S_\mathrm{final}$ with a maximum deflection angle of 15° and a resonance frequency of 100 Hz, normalized by parameter $\kappa$, see Eq.~(S12), stemming from the MOWS at three different offset angles $\uptheta_0$.
    }
\end{figure*}

Fig.~\ref{fig:1}(c) shows a time sequence of an experimental measurement.
Sinusoidal currents $I_\mathrm{coil}$ are applied to each excitation coil at the MOWS resonance frequency with orientation-dependent amplitude and sign differences, which results in an asymptotic increase of the deflection angle $\uptheta_t$ up to $\uptheta_{t,\mathrm{max}}$.
Due to a limited measurement range of the magnetometers, periodic saturation of the signals $S_1$ and $S_2$ occur.
After the excitation coils are turned off, the deflection angle decays according to Eq.~(S2) and two magnetometers measure the magnetic signals according to Eq.~(S8).
A resulting differential signal is shown in Fig.~\ref{fig:1}(d) along with a fitted curve according to the following model.

The experimental system from a top view is shown in Fig.~\ref{fig:2}(a).
It comprises the MOWS, two magnetometers and two coils for excitation of the MOWS at any in-plane orientation angle $\uptheta_0$.
Here, we define the time-dependent, total angle of the magnetic moment in the global reference frame as $\uptheta = \uptheta_0 + \uptheta_t$.
While a single magnetometer is theoretically sufficient to measure the resonance frequency of the signal, a second magnetometer with signal $S_2$, parallel to the first one with spacing $d$, is utilized to minimize the environmental noise by subtracting both signals from each other.
The final signal $S_\mathrm{final}$ for the arrangement shown in Fig.~\ref{fig:2}(a) is derived in Eqs.~(S8-S12) which reveal a complex shape of nested trigonometric functions, and hence, special care has to be taken when processing the signal.

Exemplary magnetic signal shapes according to Eq.~(S12) are shown in Fig.~\ref{fig:2}(b) for three in-plane orientation angles.
When the MOWS points towards or away from the magnetometers, i.e.~$\uptheta_0=0$° or 180°, the amplitude of the signal is maximized by the dynamic change of the magnetic dipole angle $\uptheta_t$ from $-\uptheta_{t,\mathrm{max}}$ to $+\uptheta_{t,\mathrm{max}}$ [red plot in Fig.~2(b)].
The amplitude $A$, normalized by the pre-factor $\kappa$ [see Eq.~(S10)], is in this case
\begin{eqnarray}\label{eq:amplitude_0}
    A_{\uptheta_0 = 0^\circ} &=& \left|\frac{-\sin(\uptheta_{t,\mathrm{max}})+\sin(-\uptheta_{t,\mathrm{max}})}{2} \right| \\ \nonumber
    &\approx& \uptheta_{t,\mathrm{max}}.
\end{eqnarray}
The further $\uptheta_0$ deviates from the aligned orientations, the lower the signal amplitude at the resonance frequency (first harmonics) and the higher double-frequency components (second harmonics) due to the approaching [green plot in Fig.~2(b)] or crossing [blue plot in Fig.~2(b)] of the sine-function peak during the dynamic oscillation.
For $\uptheta_0 = 90$° or 270°, due to signal symmetry, the amplitude must only be regarded from $0$° to $+\uptheta_{t,\mathrm{max}}$:
\begin{eqnarray}\label{eq:amplitude_90}
    A_{\uptheta_0 = 90^\circ} &=& \left|\frac{-\sin(90^\circ + 0^\circ)+\sin(90^\circ-\uptheta_{t,\mathrm{max}})}{2}\right| \\ \nonumber 
    &\approx& \frac{\uptheta_{t,\mathrm{max}}^2}{2}.
\end{eqnarray}
This automatically implies a decrease of the signal-to-noise ratio (SNR) for $\uptheta_0$ near 90° and 270°.
For a maximum deflection angle of the cantilever $\uptheta_{t,\mathrm{max}}$ of 15° (see supplementary materials for experimental determination of the value), the amplitude at $\uptheta_0 = 0$°, and therefore the SNR, is 7.6 times higher than the amplitude at $\uptheta_0 = 90$°.
The prominent effect on the signal fit quality $\mathrm{R}^2$ arising from this is presented in Fig.~S3.
\begin{figure*}
    \centering
    \includegraphics{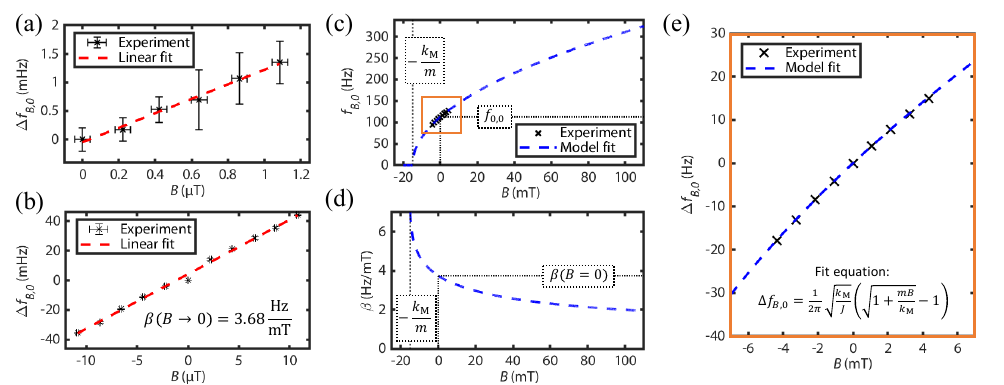}
    \caption{\label{fig:3}
    (a) Magnetic field $B$ versus resonance frequency shift $\Delta f_{B,0}$ for sub-$\upmu$T magnetic fields applied in intrinsic $x'$-direction.
    The red line shows the linear correlation between both parameters.
    (b) $B$ versus $\Delta f_{B,0}$ for $\upmu$T-range magnetic fields.
    (c) $B$ versus resonance frequency $f_{B,0}$ for mT-range magnetic fields with the physical model fit according to Eq.~(\ref{eq:delta_f}) without gradient contributions.
    Black lines indicate limit values of the model.
    (d) $B$ versus the theoretical field sensitivity $\upbeta$ according to Eq.~(\ref{eq:beta}).
    (e) Enlarged plot of $B$ versus $\Delta f_{B,0}$ for mT-range magnetic fields.
    }
\end{figure*}

The frequency shift of the oscillating magnet from an external magnetic field with magnitude $B$, pointing in $x'$-direction (Fig.~\ref{fig:1}), can be modelled as a virtual magnetic field spring constant $k_B$ parallel to the mechanical cantilever spring $k_\mathrm{M}$, with
\begin{equation}\label{eq:k_B}
    k_B = m B\cos\left(\uptheta\right) \approx m B.
\end{equation}
When the magnet, additionally, has an offset $l$ from the rotation axis, i.e.~the lever arm length shown in Fig.~\ref{fig:1}(b), an external magnetic gradient $G=\frac{\partial B_{x'}}{\partial x'}$ in the $x'$-direction results in a further, virtual gradient spring $k_G$:
\begin{equation}\label{eq:k_G}
    k_G = lmG \cos\left(2\uptheta\right) \approx lm G.
\end{equation}
The total spring constant $k$ is therefore the sum of all parallel springs, as shown in Eq.~(\ref{eq:k}).
A thorough derivation of these relationships can be found in supplementary materials [Eqs.~(S13) to (S17) and Eqs.~(S19) to (S23)].
Eqs.~(1), (4) and (5) can be generally applied to all magneto-oscillatory systems which exhibit small-angle in-plane oscillation of a magnetic moment $m>0$, and sensitivity to magnetic gradients only occurs for an oscillation offset length $l\neq0$.
The tuneability of $k_B$ by an external magnetic field can also be utilized to reduce mechanical damping contributions by a adding a second stationary magnet to the probe\cite{gleich2023}.

The resonance frequency shift $\Delta f_{B,G}$, defined as the difference of the frequency $f_{B,G}$, with magnetic field $B$ and gradient $G$, and without field or gradient $f_{0,0}$, is therefore:
\begin{equation}\label{eq:delta_f}
    \Delta f_{B,G} = f_{B,G} - f_{0,0} = f_{0,0} \left(\sqrt{1+\frac{k_B+k_G}{k_\mathrm{M}}}-1\right).
\end{equation}
This implies that $B$ and $G$ cannot be \textcolor{red}{independently} quantified as their contributions to the frequency superimpose.
We define the sensitivity of the frequency to the applied magnetic field by
\begin{equation}\label{eq:beta}
      \upbeta := \frac{\Delta f_{B,0}}{B}. 
\end{equation}
and to the magnetic gradient by
\begin{equation}\label{eq:gamma}
      \upgamma :=  \frac{\Delta f_{0,G}}{G}. 
\end{equation}
Eqs.~(\ref{eq:k_B})-(\ref{eq:gamma}) show that the frequency shifts and sensitivities are not constant for magnetic influences but instead exhibits non-linear behavior, which will be explored experimentally.

For all experimental demonstrations, we used a single 3$\times$4$\times$4 mm³ cantilever-based MOWS with a resonance frequency near $f_{0,0}=112.5$ Hz and damping $\updelta \approx 1.8$ 1/s.
The damping is neglected in the calculations of the frequency, since its effect according to Eq.~(S1) is in the sub-mHz range.
The fabrication of the device is described in supplementary materials. 

Fig.~\ref{fig:3}(a) demonstrates the frequency-sensitivity for magnetic fields below 1 $\upmu$T.
The average precision (i.e.~standard deviation) amounts to 0.33 mHz, which corresponds to $\approx$260 nT, while the wired reference magnetometer obtains a magnetic field precision of 43 nT in the magnetically noisy laboratory environment.
A linear trend is observed for very delicate frequency shifts by averaging; however, the mechanical wear of the system stemming from manually assembled components can affect the sensitivity slope.
Fabrication using MEMS technology could potentially result in even higher precision and less mechanical wear.
A magnetic field range of one magnitude larger is shown in Fig.~\ref{fig:3}(b).
Here, the sensitivity $\upbeta$ amounts to 3.68 Hz/mT from a perfectly linear relation between frequency and magnetic field.
Similarly, for the magnetic gradient of $\pm$ 50 mT/m as shown in Fig.~S4, the sensitivity $\upgamma$ is determined to 14.14 Hz$\cdot$m/T with a resolution of 65 $\upmu$T/m. 

Several orders of magnitudes larger fields in the mT-range, shown in Figs.~\ref{fig:3}(c)-(e), result in increasing non-linearity due to the square-root relation of Eq.~(\ref{eq:delta_f}).
In Fig.~\ref{fig:3}(c) the model for the frequency shift is fitted to experimental data, which is enlarged in Fig.~\ref{fig:3}(e), and extrapolated for even larger fields.
It can be seen that frequency shifts of $\pm$20 Hz for magnetic fields of $\pm$5 mT are well tolerable for the MOWS, and the frequency increases with positive fields, while it decreases with negative fields, and reaches zero when both spring constants cancel each other out, i.e.~when $k_B=-k_\mathrm{M}$ or $B=-k_\mathrm{M}/m$.
The theoretical behavior of the sensitivity $\upbeta$ over a very broad range of magnetic fields is shown in Fig.~\ref{fig:3}(d).
As $B$ decreases towards the point of spring cancellation, the sensitivity $\upbeta$ drastically increases, while for larger $B$ it diminishes, further revealing the complex asymmetric frequency shift behavior.
\begin{figure*}
    \centering
    \includegraphics{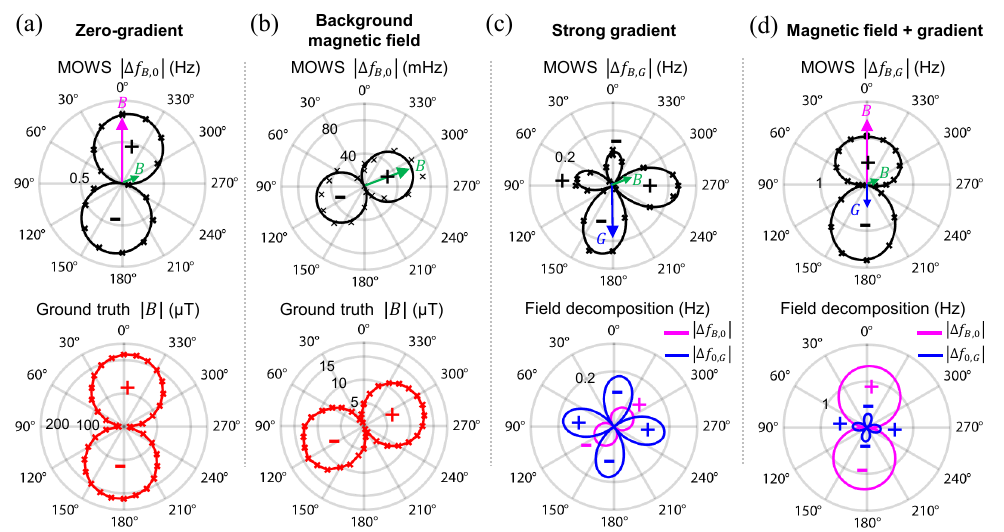}
    \caption{\label{fig:4} 
    Polar plots for mapping of parameters at various conditions by rotations about the $z$-axis by angle $\uptheta_0$.
    Only absolute values are plotted for readability and the actual corresponding signs are labeled in the plot.
    Linear versions of the plots are shown in Fig.~S5.
    (a) Frequency shift $\Delta f_{B,0}$ in a homogeneous magnetic field applied in $x$-direction (purple arrow) without gradient, with a sine fit curve and a separate ground truth measurement (red) using a wired magnetometer at the same location.
    The green arrow indicates the environmental magnetic field.
    (b) Frequency shift $\Delta f_{B,0}$ in the environmental magnetic field, with sine fit curve and ground truth measurement.
    (c) Frequency shift $\Delta f_{B,G}$ in a near-zero magnetic field with a strong gradient applied in $-x$-direction, with fit curve Eq.~(\ref{eq:decompose_fit}), and the corresponding decomposition for the magnetic field $B$ and gradient $G$.
    (d) Frequency shift $\Delta f_{B,G}$, induced by a magnetic source, with fit curve and the corresponding decomposition.
    }
\end{figure*}

Even though $\upbeta$ is undefined for $B=0$, we can use L'Hôpital's rule to obtain an essential relationship:  
\begin{equation}\label{eq:hospital}
    \lim \limits_{B \to 0} \upbeta=\frac{f_{0,0} m}{2 k_\mathrm{M}}.
\end{equation}
All parameters besides $k_\mathrm{M}$ are known or can obtained from experiments; therefore, the sensitivity for weak magnetic fields, where $\upbeta$ is in a linear regime, allows a calculation of the mechanical spring constant.
No further information about the mechanical system, such as cantilever dimensions or elastic modulus, are required to establish $k_\mathrm{M}$, and all known parameters ($\upbeta$, $f_{0,0}$, $m$) have effectively negligible errors.
Similarly, for the moment of inertia $J$ of the magnet:
\begin{equation}
    J = \frac{m}{8 \uppi^2 \upbeta f_{0,0}}.
\end{equation}
Again, no details on the cantilever length $l$ or magnetic mass $a$ with optical measurements or weighing are required to gain insight on the system.
For the presented MOWS, $k_\mathrm{M}$ and $J$ amount to $1.36\times 10^{-5}$ Nm and $2.72\times 10^{-11}$ kg m², respectively.
These values lie perfectly within the potential ranges from measured or estimated physical parameters (see supplementary materials).
This information could be, for example, used to determine absolute changes of the spring constant, or moment of inertia, by other physical influences or to examine the mechanical coupling of the MOWS to the surrounding environment \cite{fischer2022,fischer2023}.

We further demonstrate the magnetic field and gradient sensing capabilities in Fig.~\ref{fig:4} by rotating the MOWS by a defined angle $\uptheta_0$ to map the magnetic field and gradient at its location.
In a homogeneous magnetic field $B$ applied in $x$-direction without a magnetic gradient $G$, a perfect sine fit between the frequency shift $\Delta f_{B,0}$ and offset angle $\uptheta_0$ is obtained, as shown by two perfect spheres in the polar plot of Fig.~\ref{fig:4}(a).
Note that the absolute frequency shift is plotted for better visualization and the corresponding signs are given within the plot.
For ground truth comparison, a wired reference magnetometer is used.
MOWS sensing and reference align very well, and the slight tilt of the dumbbell shape away from $\uptheta_0 =$ 0°, in which the magnetic field is applied, can be attributed the superimposed environmental magnetic field, which is independently mapped in Fig.~\ref{fig:4}(b).
Minor angle differences between the MOWS and reference likely stem from probe misalignment.

Besides being able to sense magnetic fields, additionally, a cantilever-based MOWS has the unique property to be affected by magnetic field gradients $G$ at the same time.
Since $k_G$ changes with twice the rotational angle while $k_B$ only changes with the one-fold rotational angle according to Eqs.~(\ref{eq:k_B}) and (\ref{eq:k_G}), it is furthermore possible to decompose any overlaying magnetic fields and gradients to their individual contributions by the relation (derivation in supplementary materials)
\begin{equation}\label{eq:decompose}
        \Delta f_{B,G} \approx \frac{mB}{2k_\mathrm{M}}\cos\left(\uptheta\right) + \frac{l m G}{2k_\mathrm{M}}\cos\left(2\uptheta \right).
\end{equation}
This solution can be fitted to experimental data in the form of 
\begin{equation}\label{eq:decompose_fit}
    \Delta f_\mathrm{fit} = P_1 \cos(\uptheta_0 + \upphi_1) + P_2 \cos(2\uptheta_0 + \upphi_2),
\end{equation}
where $P_1$ and $P_2$ represent the prefactors, and $\upphi_1$ and $\upphi_2$ represent the global orientation of the corresponding contribution.

Fig.~\ref{fig:4}(c) shows a polar map for a near-zero magnetic field with a strong gradient in $-x$-direction.
For the gradient contribution (blue) a cloverleaf shape is obtained and a minor dumbbell shape (purple) is found stemming from the environmental magnetic fields and non-ideal centering in the anti-Helmholtz system.
The pure gradient induces a frequency shift of -183 mHz in $x$-direction.
Since $J$ is known from the aforementioned magnetic field calibration, and the magnet's mass $a$ is known, the cantilever length $l$ of Eq.~\ref{eq:k_G} amounts to 1.9 mm, which fits well in the optically estimated range (see supplementary materials).
According to Eq.~(S24), $G$ amounts to -20.8 mT/m while a reference measurement determined -11.5 mT/m.
The values are in the same order of magnitude, and the differences could arise from misalignment of the MOWS or reference magnetometer.
Since magnetic field sources generate magnetic fields and magnetic gradients a distorted dumbbell shape in Fig.~\ref{fig:4}(d) is obtained which can be decomposed to show its independent field and gradient contributions.

In summary, we present a wireless miniature sensor based on magneto-oscillatory mechanics to determine magnetic fields and gradients by measuring the oscillators frequency shift.
Rotation of the probe, furthermore, allows decomposition of the individual field and gradient contributions, and from the field sensitivity it is possible to estimate absolute values of the mechanical spring and moment of inertia. 
The analytical models can be applied to other magneto-oscillatory systems which are exposed to magnetic fields or gradients.

\section*{Supplementary material}

The supplementary material comprises details on the physical approximations of the system, thorough derivations of the presented equations, and further information on materials and methods.

\section*{Acknowledgments}
This work was partially funded by the European Union (ERC, VIBEBOT, 101041975), the MWK-BW (Az.~33-7542.2-9-47.10/42/2) and the German Cancer Research Center (DKFZ).
M.~Jeong acknowledges the support by the Stuttgart Center for Simulation Science (SimTech).

\section*{Author declarations}
\subsection*{Conflict of interests}
F.F.~and T.Q.~have a patent pending on the miniature magneto-oscillatory device (PCT/EP2023/072144).

\subsection*{Author contributions}
\textbf{Felix Fischer}: Conceptualization (equal), Data curation (lead), Formal analysis (lead), Investigation (lead), Methodology (equal), Validation (lead), Visualization (lead), Writing – original draft (lead)
\textbf{Moonkwang Jeong}: Methodology (supporting), Validation (supporting), Writing – review \& editing (supporting)
\textbf{Tian Qiu}: Conceptualization (equal), Funding acquisition (lead), Methodology (equal), Project administration (lead), Resources (lead), Supervision (lead), Validation (supporting), Writing – review \& editing (lead).

\section*{Data Availability}
The data that supports the findings of this study are available within the article and its supplementary material.


%
%

%


\bibliography{manuscript}

\end{document}


\title{Miniature magneto-oscillatory wireless sensor for magnetic field and gradient measurements} 

\author{F. Fischer}
\affiliation{Division of Smart Technologies for Tumor Therapy, German Cancer Research Center (DKFZ) Site Dresden, 01307 Dresden, Germany}
\affiliation{Faculty of Engineering Sciences, University of Heidelberg, 69120 Heidelberg, Germany}

\author{M. Jeong}
\affiliation{Institute for Physical Chemistry, University of Stuttgart, 70569 Stuttgart, Germany}

\author{T. Qiu}
\email[]{tian.qiu@dkfz.de}
\affiliation{Division of Smart Technologies for Tumor Therapy, German Cancer Research Center (DKFZ) Site Dresden, 01307 Dresden, Germany}
\affiliation{Faculty of Medicine Carl Gustav Carus, Dresden University of Technology, 01307 Dresden, Germany}
\affiliation{Faculty of Electrical and Computer Engineering, Dresden University of Technology, 01187 Dresden, Germany}

\date{\today}

\maketitle 

\renewcommand{\thefigure}{S\arabic{figure}}
\renewcommand{\theequation}{S\arabic{equation}}

\section*{Supplementary Materials}
\section*{Torsional oscillator model}
\begin{figure}[h]
    \centering
    \includegraphics{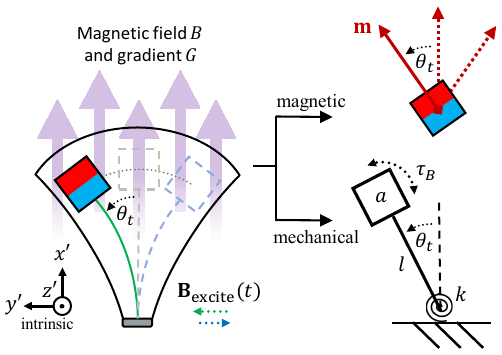}
    \caption{\label{fig:S1} 
    Approximations of the physical sub-systems. The magnetic part is approximated as pure rotation of the magnetic moment $\mathbf{m}$ and the mechanical part is approximated as torsional oscillator with mass $a$, a rigid beam of length $l$, a torsional spring constant $k$ and an additional torque $\mathbf{\uptau}_B$ on $a$ from magnetic coupling. 
    }
\end{figure}

The angular resonance frequency of a torsional oscillator, see Fig.~\ref{fig:S1}, with a rigid beam of length $l$, an oscillating mass $a$ and a torsional spring constant $k$ and damping coefficient $\updelta$ is well-defined by
\begin{equation}\label{eq:f_res}
    \upomega = \sqrt{\frac{k}{J} - \updelta^2},
\end{equation}
where $\upomega_0$ is the natural frequency and $J= a\cdot l^2$ denotes the moment of inertia of the point mass.
The deflection angle $\uptheta_t$ over time $t$ is described by an underdamped harmonic oscillator with maximum deflection angle $\uptheta_{t,\mathrm{max}}$, resonance frequency $\upomega$ and phase shift $\upphi$ $^{14}$:
\begin{equation}\label{eq:angle}
	\uptheta_t =\uptheta_{t,\mathrm{max}}  \cos\left(\upomega  t + \upphi \right)\cdot e^{-\updelta t}.
 \end{equation}
The torque $\boldsymbol{\uptau}_B$ applied on the mechanically coupled dipole with magnetic moment $\mathbf{m}$ by an external magnetic field $\mathbf{B}$ is
\begin{equation}\label{eq:torque}
    \boldsymbol{\uptau}_B = \mathbf{m}\times\mathbf{B},
\end{equation}
while the force $\mathbf{F}$ acting on the dipole, assuming that $\mathbf{m}$ is fixed in magnitude and direction, is 
\begin{eqnarray}\label{eq:force}
    \mathbf{F} &=& \nabla\left(\mathbf{m} \cdot \mathbf{B}\right) \\ \nonumber
    &=& \left(\begin{array}{c} m_{x'} \frac{\partial B_{x'}}{\partial {x'}} + m_{y'} \frac{\partial B_{y'}}{\partial x'} +m_{z'} \frac{\partial B_{z'}}{\partial {x'}} \\
    m_{x'} \frac{\partial B_{x'}}{\partial {y'}} + m_{y'} \frac{\partial B_{y'}}{\partial {y'}} +m_{z'} \frac{\partial B_{z'}}{\partial {y'}}\\
    m_{x'} \frac{\partial B_{x'}}{\partial {z'}} + m_{y'} \frac{\partial B_{y'}}{\partial {z'}} +m_{z'} \frac{\partial B_{z'}}{\partial {z'}}\end{array}\right).
\end{eqnarray}

\section*{Magnetic dipole model and signal evaluation}
\begin{figure}[h]
    \centering
    \includegraphics{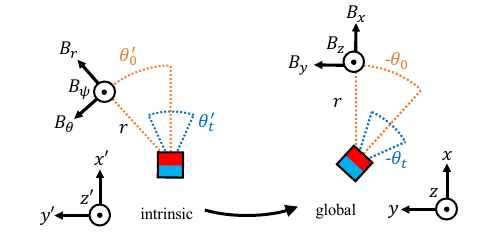}
    \caption{\label{fig:S2} 
    Coordinate transformation from intrinsic coordinates with the magnetic field in spherical coordinates to global coordinates with the magnetic field aligned to spherical coordinates, where the deflection angles $\uptheta_0$ and $\uptheta_t$ are inverted. 
    }
\end{figure}

The permanent magnet is approximated by a dipole with magnetic moment $\mathbf{m}$ in the intrinsic right-handed coordinate system $(x',y',z')$, see Fig.~\ref{fig:S1}, as
\begin{equation}\label{eq:m}
    \mathbf{m} = |\mathbf{m}|\cdot\hat{\mathbf{m}} = \frac{B_\mathrm{r}V}{\upmu_0} \cdot\left(\begin{array}{c} \cos\left(\uptheta_t\right)\\ \sin\left(\uptheta_t\right)\\0\end{array}\right)
\end{equation}
with remanence field $B_\mathrm{r}$, magnetic volume $V$ and permeability of free space $\upmu_0$.
$|\mathbf{m}|:=m$ describes the magnitude of the magnetic moment, while $\hat{\mathbf{m}}$ denotes the normalized direction vector with $\uptheta_t$ being the deflection angle from the $x'$-axis which is aligned with the magnetic moment $\mathbf{m}$ at rest.
The magnetic field $\mathbf{B}$ at distance $r$ for a dipole is calculated by using spherical coordinates ($r$,$\uptheta$,$\uppsi$) as
\begin{equation}
    \mathbf{B}_\mathrm{sph} = \left(\begin{array}{c} B_{r}\\B_{\uptheta}\\B_{\uppsi}\end{array}\right) = \frac{\upmu_0}{4\uppi}\frac{m}{r^3}\cdot\left(\begin{array}{c} 2\cos\left(\uptheta \right)\\ \sin\left(\uptheta\right)\\0\end{array}\right),
\end{equation}
where $\uptheta=\uptheta_0+\uptheta_t$ is the total angle and $\uptheta_0$ is a time-independent angular offset.
Since the magnet is subject to rotation, it is convenient to align the spherical coordinate vectors $(B_{r},B_{\uptheta},B_{\uppsi})$ to a global coordinate system $(x,y,z)$ as shown in Fig.~\ref{fig:S2}.
This results in an inversion of the total angle $\uptheta$, and by using trigonometric symmetries we obtain
\begin{equation}
    \mathbf{B}_{xyz} = \left(\begin{array}{c} B_{x}\\B_{y}\\B_{z}\end{array}\right) = \frac{\upmu_0}{4\uppi}\frac{m}{r^3}\cdot\left(\begin{array}{c} 2\cos\left(\uptheta\right)\\ -\sin\left(\uptheta\right)\\0\end{array}\right).
\end{equation}
The signal $S$ of an uniaxial magnetometer pointing in arbitrary direction $(a,b,c)$ in the  global system is then
\begin{equation}\label{eq:signal_full}
    S = \frac{\upmu_0}{4\uppi}\frac{m}{r^3} \left[a\cdot2\cos\left(\uptheta\right) - b\cdot\sin\left(\uptheta\right)\right] + N(t)
\end{equation}
with $N(t)$ as environmental noise.
Under assumption of two parallel magnetometers with signals $S_1$ and $S_2$ on a virtual line with the MOWS, and no spatial noise gradient between the sensors, the subtraction of both signals yields the same solution as Eq.~\ref{eq:signal_full}, however without noise $N(t)$ and $r$ as an effective distance
\begin{equation}
  r_{\mathrm{eff}} = \left(\frac{1}{r^3}-\frac{1}{(r+d)^3}\right)^{-1/3}. 
\end{equation}
In reality, the noise is directly dependent on the magnetometer-spacing $d$, where a lower $d$ decreases the noise, however, it simultaneously increases $r_{\mathrm{eff}}$ dramatically.

Depending on the desired application the MOWS, for example as a fixed magnetic field sensor, or with a rotational degree-of-freedom, different fitting equations are required.
Since $m$, $r$ and $d$ are known parameters from the setup and MOWS-internal magnet, we summarize the prefactor of Eq.~\ref{eq:signal_full} as
\begin{equation}
    \kappa = \frac{\upmu_0}{4\uppi}\frac{m}{r_{\mathrm{eff}}^3}
\end{equation}
and we obtain a general fit equation in the global reference frame ($\uptheta = \uptheta_0 + \uptheta_t$, see Fig.~2(a)) 
\begin{eqnarray}\label{eq:fit_general}
    S_\mathrm{fit}&=& a\cdot 2\kappa\cos\left(\uptheta_0 + \uptheta_{t,\mathrm{max}}  \cos\left(2\uppi f t + \upphi\right)\cdot e^{-\updelta t} \right) \nonumber \\ && - b\cdot \kappa\sin\left(\uptheta_0 + \uptheta_{t,\mathrm{max}}  \cos\left(2\uppi f  t + \upphi\right)\cdot e^{-\updelta t} \right) \nonumber \\ && + B_\mathrm{DC}
\end{eqnarray}
where $B_\mathrm{DC}$ is a constant offset due to environmental magnetic fields.

In all presented experiments, we fix the magnetometer orientation to $b=1$, see Fig.~2(a), which yields
\begin{equation}\label{eq:fit}
    S=-\kappa\sin\left(\uptheta_0 + \uptheta_{t,\mathrm{max}}  \cos\left(2\uppi f  t + \upphi\right)\cdot e^{-\updelta t} \right) + B_\mathrm{DC}
\end{equation}
as the full fit equation used for signal evaluation.
Furthermore, in the special arrangement of $\uptheta_0=0$, it is possible to further deduce $\uptheta_{t,\mathrm{max}}$ for calibration, which is constant when the cantilever is excited to its physical limit position due to the housing.

\begin{figure}[h]
    \centering
    \includegraphics{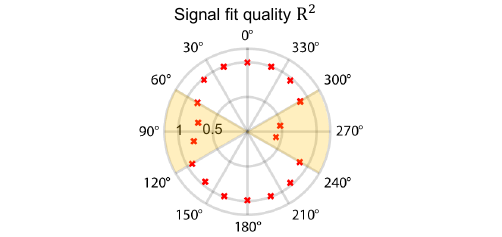}
    \caption{\label{fig:S3} 
    Signal fit quality $\mathrm{R}^2$ versus offset angle $\uptheta_0$ for a full rotation of the MOWS.
    Shaded areas indicate a reduced fit quality due a decreasing signal amplitude.
    }
\end{figure}

The signal-to-noise ratio or signal fit quality $\mathrm{R}^2$ of Eq.~(6) to the real signal, as explained in the main text, is highly dependent on the angular offset $\uptheta_0$.
This can be seen in Fig.~\ref{fig:S3} for a full rotation of the MOWS, taken from the measurement of Fig.~4(a) in a homogeneous magnetic field without gradient. 
In a range of approximately $\pm$ 60° around the $\uptheta_0 =$0° and 180° orientations, the signal fit quality is very high, reaching values up to 0.997 at 0° and an average of 0.985.
Outside of these regions, indicated as shaded areas in Fig.~\ref{fig:S3}, $\mathrm{R}^2$ drops down to a minimum of 0.423 at 260° or an average value of 0.766.
Hence, if only a single magnetic field or gradient component needs to be measured, i.e.~the MOWS does not need to be rotated, it is always beneficial to choose the alignment with $\uptheta_0 = 0$° or 180° to maximize the signal strength. 

\section*{Magnetic field spring constant}
The underlying reason for the magneto-mechanical interaction is stemming from the torque $\boldsymbol{\uptau}_B$ in Eq.~(\ref{eq:torque}) which is applied on the magnet and will be explained in the following.
A generalized solution of Eq.~(\ref{eq:torque}) for an arbitrary external field $\mathbf{B}$ using the magnetic moment of Eq.~(\ref{eq:m}) is
\begin{equation}
    \boldsymbol{\uptau}_B = m B \cdot\left(\begin{array}{c}\sin\left(\uptheta_t\right)\hat{B}_{z'}\\ -\cos\left(\uptheta_t\right)\hat{B}_{z'}\\ \cos\left(\uptheta_t\right)\hat{B}_{y'}-\sin\left(\uptheta_t\right)\hat{B}_{x'}\end{array}\right).
\end{equation}
Torque components in $x'$- and $y'$-direction theoretically result in a rotation of the magnet about the respective axes, however, due to the rectangular shape of the cantilever cross-section, the area moment of inertia are highly unfavorable for such rotations and generate negligible deflections.
Only the torque components in $z'$-direction results in noticeable deflections in the $x'$-$y'$-plane, and  only $x'$- and $y'$-components of the magnetic field are of further relevance.
Therefore, the magnitude of the torque $|\boldsymbol{\uptau}_B|=\uptau_{B,z'}$ for a magnetic field magnitude $B$, in the $x'$-$y'$-plane, is
\begin{eqnarray}
     \uptau_{B,z'} &=& m B\cdot \left(\cos\left(\uptheta_t\right)\hat{B}_{y'}-\sin\left(\uptheta_t\right)\hat{B}_{x'}\right)\nonumber\\&\approx&  m B_{y'} - m B_{x'}\cdot\uptheta_t\nonumber\\&:=& \uptau_{B,\mathrm{stat}} + \uptau_{B,\mathrm{dyn}},   
\end{eqnarray}
where the middle step is a small angle approximation.
At resting position of the magnet $\uptheta_t=0$°, $\uptau_B$ is constant, only depending on the magnetic field perpendicular to the magnetic moment axis.
This torque results in a static, time-independent contribution $\uptau_{B,\mathrm{stat}}$ to the deflection angle.
Only the latter term , which comprises the magnetic field in the main direction $x'$, is time-dependent and therefore affects the dynamic behavior.

During a periodic oscillation, the dynamic torque $\uptau_{B,\mathrm{dyn}}$ reaches opposing signs.
For $B_{x'}>0$, $\uptau_{B,\mathrm{dyn}}$ is negative for positive deflection angles and positive for negative deflection angles.
Since the torque results in a direct force on the magnet due to the assumption of a rigid connection (Fig.~\ref{fig:S1}), positive and negative deflection angles result in a force towards the resting position.
The deflection is therefore increasing the resonance frequency, and for $B_{x'}<0$, the effect is inverted, decreasing the resonance frequency.
This principle can be described by a virtual magnetic torsional spring with spring constant $k_B$ which is in parallel to the mechanical spring with $k_\mathrm{M}$.
The overall spring constant $k$ is calculated by
\begin{equation}
    k = k_\mathrm{M} + k_B,
\end{equation}
where $k_B$ is not limited to only positive values like the mechanical spring constant, and
\begin{equation}\label{eq:k_B}
    k_B = -\frac{\mathrm{d} \uptau_{B,\mathrm{dyn}}}{\mathrm{d} \uptheta_t} = m B_{x'}\cdot\cos\left(\uptheta_t\right)\approx m B_{x'},
\end{equation}
or in global coordinates, see Fig.~\ref{fig:S2},
\begin{equation}\label{eq:k_B_global}
    k_B = m B_{x}\cdot\cos\left(\uptheta_0 + \uptheta_t\right).
\end{equation}
Note, that the minus sign disappears due to Hooke's law for torsional springs $\uptau = -k\uptheta$.
The resonance frequency within the magnetic field according to Eq.~(\ref{eq:f_res}) with negligible damping therefore becomes
\begin{equation}\label{eq:omega_res_B}
    \upomega_\mathrm{B} = \sqrt{\frac{k_\mathrm{M} + k_B}{J}}.
\end{equation}

\section*{Estimation of resonator properties}
For the moment of inertia $J=a\cdot l^2$, the magnet's mass $a$ was measured to be 7.2 $\times 10^{-6}$ kg.
The cantilever length $l$ cannot be accurately determined and is estimated to be between 1.5 mm and 2.5 mm.
Accordingly, $J$ is between 1.62 $\times 10^{-11}$  kg m² and $4.5 \times 10^{-11}$ kg m².
Using the relationship of Eq.~\ref{eq:f_res} with $f=112.48$ Hz, we obtain a mechanical spring constant between $0.81\times 10^{-5}$ Nm and $ 2.25 \times 10^{-5}$ Nm.
These values are used as physically possible ranges.

\section*{Magnetic gradient spring constant}
\begin{figure}[h]
    \centering
    \includegraphics{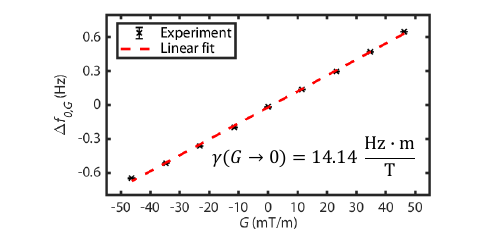}
    \caption{\label{fig:S4} 
    Magnetic gradient $G$ versus resonance frequency shift $\Delta f_{0,G}$.
    The linear slope corresponds the the gradient sensitivity $\upgamma$.
    }
\end{figure}

Here, we again operate in the intrinsic coordinate system where we assume that the magnetic moment $\mathbf{m}$ points in $x'$ direction (see Fig.~\ref{fig:S2}). Now, we apply exclusively in this direction a finite homogeneous magnetic gradient, $\frac{\partial B_{x'}}{\partial {x'}} = G_{x'}$, meaning that all other spatial gradients for all other magnetic field directions are zero.
These assumptions, according to Eq.~\ref{eq:force}, result in a force acting in $x'$-direction:
\begin{eqnarray}
    \mathbf{F} &=& \left(\begin{array}{c} m_{x'} G_{x'} \\0\\ 0\end{array}\right) \\ \nonumber
    F_{x'} &=& m G_{x'} \cos\left(\uptheta_t \right),
\end{eqnarray}
with $m = |\mathbf{m}|$ and $\uptheta_t$ as the time-dependent deflection angle.
Again, as shown in Fig.~\ref{fig:S1}, we assume a rigid cantilever of length $l$ with a torsional spring, and calculate the gradient-based torque $\boldsymbol{\uptau}_\mathrm{G}$ acting on the base in 3D as
\begin{equation}
    \boldsymbol{\uptau}_G = \mathbf{l} \times \mathbf{F}.
\end{equation}
The only favorable area moment of inertia is in the $z'$-direction which yields
\begin{equation}
    \uptau_{G,z'} = -l_{y'} F_{x'} = -l \sin\left(\uptheta_t \right) m G_{x'} \cos\left(\uptheta_t \right) 
\end{equation}
with $l_{y'} = l \sin\left(\uptheta_t \right)$ as the $y'$-component of the lever.
The magnetic gradient spring constant $k_G$, calculated similar to Eq.~(\ref{eq:k_B}), is then
\begin{equation}
    k_G = -\frac{\mathrm{d} \uptau_{G,z'}}{\mathrm{d} \uptheta_t} =l m G_{x'} \cos\left(2\uptheta_t \right) \approx l m G_{x'}
\end{equation}
or in global coordinates
\begin{equation}\label{eq:k_G_global}
    k_G = l m G_{x} \cos\left(2\uptheta_0 + 2\uptheta_t \right).
\end{equation}
While the force on the magnet scales with the cosine of the angle, the lever for torque scales with a sine, which leads to maxima of the spring constant at offset angles $\uptheta_0$ of 0° and 180°, and minima at 90° and 270°.
By rearranging Eq.~(\ref{eq:k_G_global}) for $G_{x}$ and replacing $k_G$ by the rearranged form of Eq.~(\ref{eq:omega_res_B}) (replace $k_B$ by $k_G$), we finally obtain
\begin{equation}
    G=\frac{4\uppi^2(f + \Delta f_{0,G})^2 J - k_\mathrm{M}}{lm}.
\end{equation}
The frequency shift by application of a magnetic gradient is shown in Fig.~\ref{fig:S4}.
The shift from residual magnetic fields in the anti-Helmholtz setup were compensated using the known field sensitivity $\upbeta$.
The resolution, as the average standard deviation of each measurement, corresponds to 65 $\upmu$T/m. 

\section*{Magnetic field and gradient decomposition}
\begin{figure*}
    \centering
    \includegraphics{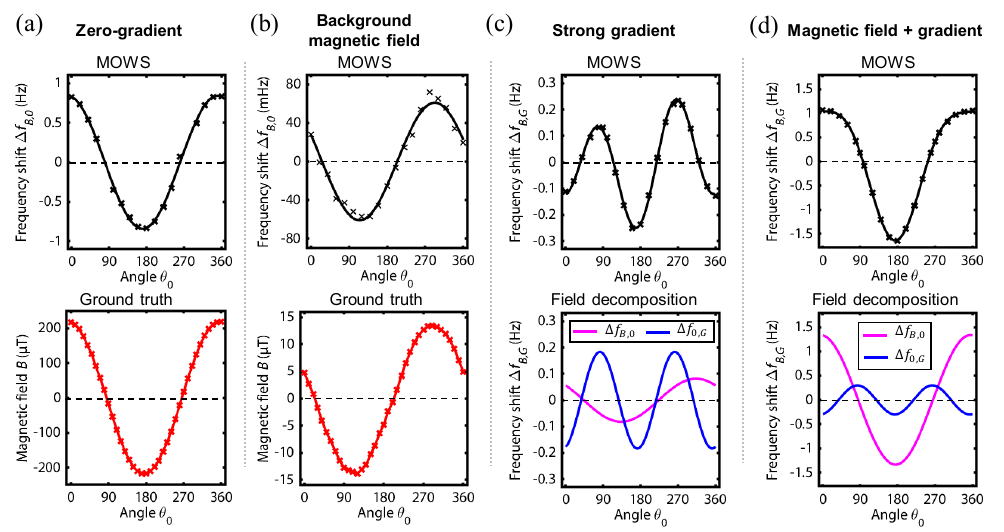}
    \caption{\label{fig:S5} 
    Linear plots for mapping of parameters at various conditions by rotations about the $z$-axis by angle $\uptheta_0$.
    Corresponding polar plots are shown in Fig.~4.
    (a) Frequency shift $\Delta f_{B,0}$ in a homogeneous magnetic field applied in $x$-direction (purple arrow) without gradient, with a sine fit curve and a separate ground truth measurement (red) using a wired magnetometer at the same location.
    (b) Frequency shift $\Delta f_{B,0}$ in the environmental magnetic field, with sine fit curve and ground truth measurement.
    (c) Frequency shift $\Delta f_{B,G}$ in a near-zero magnetic field with a strong gradient applied in $-x$-direction, with fit curve Eq.~(\ref{eq:decompose_fit_S}), and the corresponding decomposition for the magnetic field $B$ and gradient $G$.
    (d) Frequency shift $\Delta f_{B,G}$, induced by a magnetic source, with fit curve and the corresponding decomposition.
    }
\end{figure*}

The total spring constant, with contributions from magnetic fields and magnetic gradients, is
\begin{equation}
    k = k_\mathrm{M} + k_B + k_G.
\end{equation}
The magnetic field spring constant $k_B$ is dependent of $\uptheta_0$ (see Eq.~(\ref{eq:k_B_global})), while the gradient spring constant $k_\mathrm{G}$ is dependent on $2\uptheta_0$ (see Eq.~(\ref{eq:k_G_global})).
The change of frequency from the magnetic spring constants is
\begin{eqnarray}
        \Delta f &=& f_{B,G} - f_{0,0} \nonumber \\  &=& f_{0,0} \left(\sqrt{1+\frac{k_B+k_G}{k_\mathrm{M}}}-1\right).
\end{eqnarray}
For small magnetic fields and gradients, i.e. $k_B+k_G<<k_\mathrm{M}$, we can use the Taylor series expansion of $\sqrt{1+u}$ for small $u$ which is $1 + u/2$ and we obtain
\begin{eqnarray}
    \Delta f &\approx& \frac{k_\mathrm{B}+k_\mathrm{G}}{2k_\mathrm{M}} \nonumber \\ &=& \frac{m B_{x}}{2k_\mathrm{M}}\cdot\cos\left(\uptheta\right) + \frac{ l m G_{x}}{2k_\mathrm{M}}\cdot  \cos\left(2\uptheta \right).
\end{eqnarray}
This solution can be fitted to experimental data in the form of 
\begin{equation}\label{eq:decompose_fit_S}
    \Delta f_\mathrm{fit} = P_1 \cos(\uptheta_0 + \phi_1) + P_2 \cos(2\uptheta_0 + \phi_2),
\end{equation}
where $\phi_1$ and $\phi_2$ represent the global orientation of the corresponding contribution.
Hence, it is possible to separate independent contributions of the magnetic field and gradient without previous knowledge of either contribution, only by rotation of the probe around its own axis.
This is shown in Fig.~\ref{fig:S5} in linear plots, corresponding to the polar plots of Fig.~4.

\section*{Materials and Methods}
\subsection*{MOWS fabrication}
The device was fabricated by a previously described process $^{14}$, which is repeated here for convenience.
The housing was 3D-modelled (Inventor Professional 2021, Autodesk, US) and printed with 50 µm resolution (3L, Formlabs, US) and translucent resin (Clear V4, Formlabs).
A cavity within the housing allows deflection of the oscillating beam and also limits the maximal deflection angle.
The oscillation frequency is tuneable by choice of cantilever dimensions and material to achieve frequencies within the range of approximately 50 Hz to 500 Hz.

For the presented MOWS, a $\sim$1.5-2.5 mm $\times$ 0.2 mm $\times$ 20 µm stripe (C1095 spring steel, Precision Brand, US) was laser-cut (MPS Advanced, Coherent, US) as the cantilever, and a $\diameter$ 1 mm $\times$ 1 mm cylindrical NdFeB magnets (N52, Guys Magnets, UK) with a theoretical magnetic moment of 0.89 mAm² and axial magnetization was attached to the end of the cantilever using cyanoacrylate adhesive (Loctite 401, Henkel, Germany).
The cantilever with attached magnet was threaded through an opening of the housing and was tightly fixed to the housing with adhesive (UHU Hart, UHU, Germany) and 30 min curing at 60°C.
A lid was finally glued onto the housing to fully enclose and protect the cantilever.
The presented MOWS exhibited a resonance frequency around 112.5 Hz with a damping of 1.8 1/s.
All measurements were conducted with the same MOWS.

\subsection*{Experimental setups and measurements}
Experiments were carried out in two variations to achieve 1) the largest spacing between MOWS and wired stationary devices, and 2) a homogeneous magnetic field without gradient and zero-field with a strong-gradient.
Both setups use two fluxgate magnetometers (FL1-10-10-AUTO, Stefan Mayer Instruments, Germany) with 0.1 nT resolution and a $\pm$10 $\upmu$T range.
Analog magnetic signals were digitized by a data acquisition board (PCIe-6363, NI, US) with a sampling rate of 20 kS/s in Python (V3.9.12, Python Software Foundation, US) and processed in MATLAB (R2022b, The MathWorks, US).

The two excitation coils (LSIP-330, Monacor, Germany) with an inductance of 3.3 mH were powered by power amplifiers (TSA 4000, the t.amp, Germany) with a current amplitude of up to 10 A.
The sinusoidal excitation current magnitudes and signs were adjusted depending on the orientation of the MOWS to ensure excitation to the maximum deflection angle by generating magnetic fields perpendicular to the magnets magnetization axis. 
The excitation frequency was adjusted twice before recording 5 independent signals with 10 s gaps in-between to avoid over-straining of the sensitive cantilever.
These 5 signals were evaluated independently and 60 periods of each signal (100 periods in Fig.~3(a) for very small frequency changes) where used to calculate the average frequency and standard deviation.
The maximum deflection angle $\uptheta_{t,\mathrm{max}}$ of the MOWS was calibrated to 24°, which could render small-angle approximations inaccurate, however, seen over 60 periods ($\approx$ 0.5 s) with the mentioned resonance frequency and damping, the average angle is 14.9°, which is below the 4\% error margins for a cosine and near 1\% error for a sine.

Ground truth measurements were recorded for 1 s using a separate fluxgate magnetometer (Fluxmaster, Stefan Mayer Instruments, Germany) with a tunable range from $\pm$100 $\upmu$T to $\pm$1 $\upmu$T.
Custom non-magnetic stages were 3D-modelled (Inventor Professional 2021, Autodesk, US) and printed (Bambu Lab X1 Carbon, BAMBULAB LIMITED, China) to achieve rotation.
External power for static magnetic fields was provided by a power supply (NGM202, Rohde \& Schwarz, Germany) with 0.1 mA resolution and a maximum current of 2 A.

\subsubsection*{Largest spacing setup}
This setup is shown in Fig.~2(a) and used for experiments shown in Fig.~4(b) and (d).
The two coils are arranged such that the air gap between the coils, the magnetometers and the MOWS is maximized to demonstrate a realistic application setting where the wired components are all within one plane.
The effective distance $r_\mathrm{eff}$ of the signal is 6.75 cm while the true spacing $r$ is 6 cm.
To generate magnetic fields perpendicular to the magnets magnetization axis, the base current amplitude (of up to 10 A) of the sinusoidal wave is multiplied by $\sin\left(\uptheta_0+135^\circ\right)$ for coil 1 and multiplied by $\sin\left(\uptheta_0+45^\circ\right)$ for coil 2.
For the experiment in Fig.~4(d), an additional coil (LSIP-330, Monacor, Germany) was placed at 5 cm distance from the MOWS in $-x$-direction and powered with a current of 500 mA.

\subsubsection*{Helmholtz/Anti-Helmholtz setup}
This setup was used for experiments shown in Fig.~3 and Fig.~4(a) and (c). 
To achieve a Helmholtz configuration, two coils (EM-6723A, Pasco, US) with a spacing and radius of 10.5 cm were placed in $x$-direction around the MOWS.
Due to spatial constraints, the coils 1 and 2 were placed at $-y$ and $-x$ from the MOWS, respectively
Correspondingly, the base current amplitude of the sinusoidal wave is multiplied by $\sin\left(\uptheta_0+90^\circ\right)$ for coil 1 and multiplied by $\sin\left(\uptheta_0+0^\circ\right)$ for coil 2.
For experiments in Fig.~3(a), (b) and (c) currents of 1 mA to 1.5 mA, -5 mA to 5 mA and -2 A to 2 A, respectively, were applied.
For experiments in Fig.~4(a), the coils where connected in parallel (Helmholtz configuration) with a current of 100 mA to achieve a homogeneous magnetic field without gradient, and connected anti-parallel (anti-Helmholtz configuration) for Fig.~4(c) with a current of 500 mA to achieve a homogeneous magnetic gradient with a near-zero field magnitude.

\bibliography{MOWSbib}